\documentclass[conference,a4paper]{IEEEtran}
\usepackage{cite}
\usepackage{amsmath,amssymb,amsfonts}
\usepackage{algorithmic}
\usepackage{graphicx}
\usepackage{textcomp}
\usepackage{xcolor}
\usepackage{multirow}
\usepackage{booktabs}
\usepackage[acronym]{glossaries}
\usepackage[colorlinks=true, urlcolor=blue, citecolor=blue, linkcolor=darkgray]{hyperref}

\def\BibTeX{{\rm B\kern-.05em{\sc i\kern-.025em b}\kern-.08em
    T\kern-.1667em\lower.7ex\hbox{E}\kern-.125emX}}
\begin{document}
\newacronym{ldm}{LDM}{Latent Diffusion Model}
\newacronym{vae}{VAE}{Variational Autoencoder}
\newacronym{absrel}{AbsRel}{Absolute Mean Relative Error}
\newacronym{rmse}{RMSE}{Root Mean Square Error}
\title{LiDAR Resolution Recovery via Foundation-Model-Guided Diffusion}

\author{
\IEEEauthorblockN{Samed Do\u{g}an, Nico Leuze, Alfred Schöttl}
\IEEEauthorblockA{Dept. of Electrical Engineering and Information Technology\\ 
Munich University of Applied Sciences, 80335 Munich, Germany\\
Email: samed.dogan@hm.edu\\}
}

\maketitle

\begin{abstract}
% ==== Samed: ABSTRACT DONE =================== %
High-beam-count LiDAR sensors are costly, yet many perception pipelines require dense angular sampling. Using a pretrained Stable Diffusion model as the backbone, we fine-tune a LiDAR-conditioned depth model with pseudo-depth targets from a 2D foundation model. During training, the LiDAR conditioning is randomly decimated at different beam budgets. We then investigate how much of a LiDAR scan can be recovered from heavily decimated input and characterize performance across the input beam budget. We evaluate against physically held-out real beams on nuScenes and report recovery separately from fit accuracy. Our model yields its largest advantage in very sparse regimes, achieving a $\delta_{1.25}$ accuracy of $66.8$\% from $4$-beam input where scattered interpolation reaches only $45.1$\%. A class-stratified error breakdown further reveals that planar surfaces recover first while objects introducing depth discontinuities degrade earliest. Together, these results quantify the recovery/resolution trade-off for foundation-model-guided LiDAR enhancement.
\end{abstract}


\section{Introduction}

Accurate 3D geometric perception is fundamental to autonomous navigation. Consequently, LiDAR sensing systems have gained widespread adoption in automotive and robotic applications \cite{application-automotive1, application-automotive2, application-robotics1}. However, high-resolution LiDAR sensors carry a steep cost in terms of energy and data rate consumption, despite being the superior choice for 3D scene understanding. Low beam-count sensors offer a realistic trade-off between overall system cost and performance. A recurring practical question is therefore whether a low-beam sensor can be used to approximate the angular fidelity of a costly one. Several recent works address LiDAR generation using diffusion-based frameworks \cite{lidar1-lidarDM, lidar2-r2dm, lidar3-lidargen, lidar5-LiDM, lidar6-rangeldm}. These models operate on spherical range images, parameterized by the LiDAR's elevation and azimuth angles. Even though this formulation is a native representation of sensor geometry, it is confined to the sensor's own measurement manifold. In contrast, models operating in camera-space can inject dense priors from 2D foundation models, including geometry, semantics, and object boundaries learned at web scale, into the recovery process \cite{dogan}. Diffusion-based LiDAR upsampling is most often cast as a masked completion or super-resolution task (Sec.~\ref{sec:completion}), typically relying on high-resolution LiDAR supervision. Building upon the architecture introduced in \cite{dogan}, our approach operates without high-resolution LiDAR supervision and recovers physically held-out beams as dense depth maps in the camera view from sparse LiDAR conditioning under 2D foundation model supervision.
Concretely, our contributions are as follows:
\begin{itemize}
    \item We report a degradation curve of recovery accuracy as a function of input beam budget, evaluated against held-out real beams.
    \item We provide a class-stratified failure analysis, inspecting where recovery breaks down first under very sparse inputs.
    \item We inspect the influence of scene prompt description on reconstruction quality, when a large portion of information is missing.
\end{itemize}

The study is scoped to camera-view depth recovery against real held-out measurements, characterizing the additional properties of image-space diffusion relative to the native spherical domain representation.
\section{Related Work}
\subsection{Generative LiDAR Models}
% ====== SUBSECTION - DONE
Motivated by the success of diffusion and score-based generative models  \cite{diffusion1-ddpm, diffusion3-earlywork, diffusion4-score}, a growing body of work models LiDAR scans directly in their native spherical range representation \cite{lidar1-lidarDM, lidar2-r2dm, lidar3-lidargen, lidar5-LiDM, lidar6-rangeldm}, extending image-domain latent diffusion \cite{diffusion2-ldm} to the range image parameterized by sensor azimuth and elevation. These models draw their prior entirely from the sensor's own measurement manifold, which preserves range statistics but ties the model to a single sensor geometry. Our setting instead conditions an image diffusion model on the projected scans and draws its prior from outside the LiDAR distribution. We therefore view our approach as complementary, trading native 360\textdegree\, coverage for access to dense 2D priors.

\subsection{LiDAR Completion and Super-Resolution}
\label{sec:completion}
% ============== I think this is done aswell ===== 
Several range-domain diffusion models address sparse scan completion, but by different mechanisms. R2DM \cite{lidar2-r2dm} performs completion at sampling time via RePaint-style inpainting \cite{lugmayr2022repaint}. RangeLDM \cite{lidar6-rangeldm} trains a conditional latent diffusion model for upsampling, mapping a sparse range image to a dense one under dense-LiDAR supervision. A separate line of work fuses sparse projected depth maps with registered camera images to regress dense metric depth \cite{depth-comp1, depth-comp2}, typically supervised by accumulated LiDAR. Lastly, \cite{gui2025depthfm} demonstrates the robustness of diffusion-based dense depth estimation under sparsity within the 2D image domain. Unlike these approaches, we require no dense-LiDAR supervision, train under a continuum of beam budgets, and measure per-beam recovery on physically held-out points in camera-view depth. This enables evaluation of camera-space depth recovery under physically decimated LiDAR conditioning rather than dense-LiDAR supervised upsampling.

\section{Method}
\subsection{Background}
\label{sec:backbone}
The camera-space projection and latent construction follow the architecture introduced in \cite{dogan}. Let $\mathcal{P}=\{\mathbf{p}_i\}_{i=1}^N$ be a LiDAR scan in the sensor frame, where each $\mathbf{p}_i\in\mathbb{R}^3$ has a corresponding intensity $r_i$. Given calibrated camera intrinsics $\mathbf{K}$ and LiDAR-to-camera extrinsics $\mathbf{T}_{CL}$, the scan is rendered into a conditioning latent $\mathbf{z}_c$ using the projection operator $\Pi\in\mathbb{R}^2$:
\begin{equation}
    \mathbf{z}_c=\Pi(\mathcal{P};\mathbf{T}_{CL}, \mathbf{K})\in \mathbb{R}^{2\times H \times W}\;.
    \label{eq:proj}
\end{equation}
Its two channels contain depth and intensity, with occlusion consistency enforced by z-buffering. The conditioning is constructed for all $V$ camera views. $\Pi$ is applied after the beam decimation described in Sec.~\ref{sec:decimation}, ensuring that the conditioning only contains projections of retained beams. The training objective uses the canonical $\epsilon$-parameterized denoising objective for \glspl{ldm} \cite{diffusion2-ldm}. The dense depth target is encoded by a frozen \gls{vae} \cite{Kingma2014} into $\mathbf{z}_0$ and corrupted with Gaussian noise to obtain $\mathbf{z}_t$ at timestep $t$. A U-Net \cite{unet} $\boldsymbol{\epsilon}_\theta$ is trained to invert this process, conditioned on the max-pooled projection $\mathbf{z}_c$ concatenated with $\mathbf{z}_t$ along the channel dimension:
\begin{equation}
    \mathcal{L}_{LDM}=\mathbb{E}_{\mathbf{z}_0,\boldsymbol{\epsilon},t}\Big[w(t)\lVert\boldsymbol{\epsilon}-\boldsymbol{\epsilon}_\theta\left([\mathbf{z}_t,\,\mathbf{z}_c], t, c\right)
    \rVert_2^2\Big],
    \label{eq:loss}
\end{equation}
where $[\cdot,\cdot]$ denotes channel concatenation, $c$ is the text conditioning and $w(t)$ is the SNR-weighted schedule \cite{hang2023efficient}.
\subsection{Randomized Beam Decimation}
\label{sec:decimation}
We treat the native $32$-beam nuScenes \cite{nuscenes} scan as the dense reference and condition the model on the decimated view. Decimation is performed on the physical LiDAR beams using their native ring indices $\rho_i\in\{0,\dots,31\}$. We refer to the number of retained beams as the beam budget $b$.
\paragraph{Budget Sampling}
% ========== DONE ============== 
The beam budget is sampled per scan as $b\sim\mathcal{U}(\{b_{\min},\dots,32\})$ with $b_{\min}=4$, rather than using fixed sparsity levels. Training across the full budget range reduces overfitting to a fixed set of elevation angles and yields a single model evaluable at every budget. The degradation curve in Sec.~\ref{sec:degradation_curve} therefore reports direct measurements at each budget, rather than interpolation between separately trained models.
\paragraph{Ring Selection}
% ============== DONE ===============
We mimic a lower-beam-count sensor by selecting LiDAR rings at a uniform stride over the available ring indices determined by the budget $b$. This yields two subsets of $\mathcal{P}$: the retained set $\mathcal{P}_R\subseteq\mathcal{P}$ that forms the conditioning given Eq.~\ref{eq:proj}, and the complement $\mathcal{P}_H=\mathcal{P}\setminus\mathcal{P}_R$, which is withheld for evaluation. The dense supervision target is independent of the beam budget. Consequently, the model is tasked to reconstruct the same dense depth from varying amounts of observed information. Furthermore, the stride phase is randomly sampled during training to prevent the model from memorizing a fixed subset of ring positions. During evaluation, we keep the stride phase fixed to zero.
\subsection{Evaluation Protocol}
\label{sec:protocol}
% ============ DONE ============
Depth recovery is evaluated on the held-out point set $\mathcal{P}_H$ of the nuScenes \cite{nuscenes} validation split. For each unseen scene, the procedure in Sec.~\ref{sec:decimation} is applied at a fixed beam budget, yielding the disjoint retained and held-out sets. This enforces separation at two levels: unseen scenes prevent scene-level memorization, while held-out beams prevent evaluation from reducing to a mere reproduction of the conditioning measurements. Dense per-view relative depth is then aligned to metric scale and shift by least-squares using only the retained points $\mathcal{P}_R$. The aligned metric depth is scored against the held-out returns at their projected pixel locations.
We report the \gls{absrel}, \gls{rmse} and $\delta_{1.25}$ metrics widely adopted in dense depth estimation literature \cite{marigold, midas, gui2025depthfm, da3}. As a baseline, the retained projection is densified using nearest-neighbor scattered interpolation over valid retained pixels. This isolates the contribution of the learned prior from geometry recoverable by interpolation alone. We skip the affine transformation for the baselines given their native metric scale. 
\section{Experiments}
\label{sec:experiments}
\subsection{Setup}
\label{sec:setup}
% ============ Needs adjustment if have continous values ==========
Training is performed on the nuScenes \cite{nuscenes} training split under randomized beam decimation (Sec.~\ref{sec:decimation}) and evaluated on the validation split. The U-Net is initialized with Stable Diffusion 1.5 weights. Dense depth pseudo-labels are generated with Depth Anything 3 \cite{da3} using the depth preprocessing procedure described in \cite{dogan}. All recovery metrics are computed on held-out beams, with per-view scale and shift fit on retained beams only (Sec.~\ref{sec:protocol}). 
We trace the degradation curve (\figurename~\ref{fig:class_curve} left) by evaluating every integer budget $b\in\{4,\dots,32\}$. We train for 100 epochs on mixed precision using AdamW with a learning rate of $5\times10^{-4}$, weight decay of $0.01$ and a minimum learning rate of $5\times10^{-5}$ on a cosine schedule with 1000 warmup steps. We use a batch size of 32 with 6 views per batch on 8 NVIDIA A40 GPUs, resulting in an effective batch size of $1536$.  
\subsection{Degradation Curve}
\label{sec:degradation_curve}
\begin{figure}
    \centering
    \includegraphics[width=0.7\linewidth]{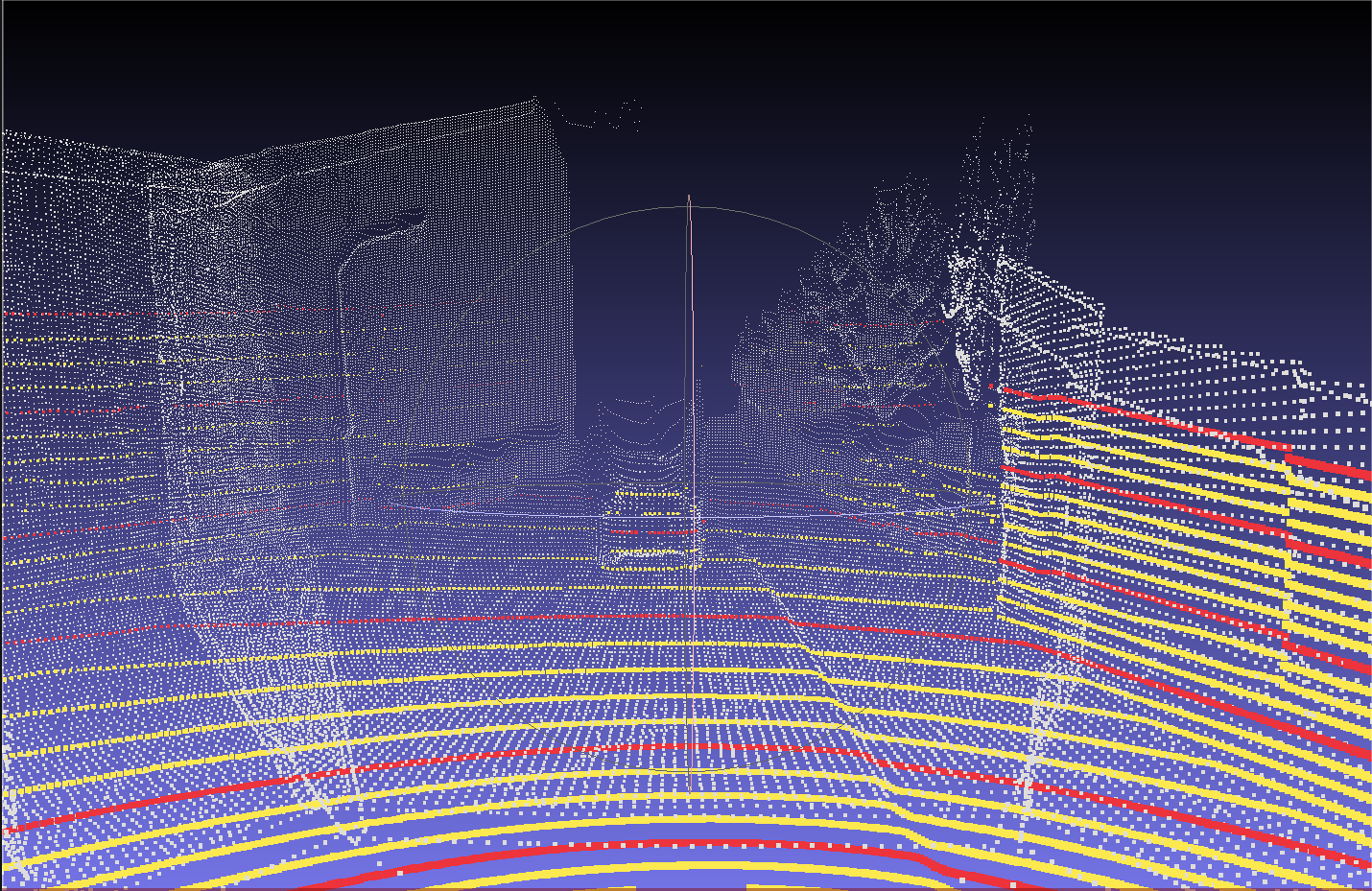}
    \caption{Partial view of the model output in point cloud space across $3$ fused views (gray) given $8$ input beams (red) compared to ground truth (yellow).}
    \label{fig:upsampled}
\end{figure}

% ======== CURVE Neeeded ==========
\tablename~\ref{tab:degradation} reports representative budgets, while the full curve appears in \figurename~\ref{fig:class_curve} (left). The proposed model outperforms interpolation at every budget, with the largest \gls{absrel} improvement in the sparsest regime, where geometry cannot be recovered by interpolation alone ($0.692$ vs. $0.260$ at $b\!=\!4$). The gap narrows as beam density increases and interpolation becomes more effective ($0.151$ vs. $0.271$ at $b\!=\!16$). Recovery error decreases monotonically as more beams are retained and plateaus at $b\!=\!16$. The full-density fit metric ($b\!=\!32$) provides a reference for reconstruction without beam withholding, yielding \gls{absrel} $0.147$. A qualitative example of geometric reconstruction with $b\!=\!8$ is given in \figurename~\ref{fig:upsampled}.
%Figure~\ref{fig:curve} plots $\delta_{1.25}$ across the continuous budget range.
\begin{table}
    \centering
    \footnotesize
    \setlength{\tabcolsep}{3pt}
    \caption{Held-out beam recovery vs. beam budget. ``Nearest'' applies nearest valid-pixel scattered interpolation. At $b\!=\!32$ no beams are held out. The fit is reported as a full-density reference.}
    \label{tab:degradation}
    \begin{tabular}{llccc}
    \toprule
    Budget & Method & \gls{absrel}$\downarrow$ & \gls{rmse}$\downarrow$&$\delta_{1.25}\uparrow$\\
    \midrule
    \multirow{3}{*}{$b{=}4$} & Nearest   & 0.692 & 12.62 & 0.451 \\
                             & Ours (no scene)   & \textbf{0.260} & \textbf{8.67} & \textbf{0.668} \\
                             & Ours (scene) & 0.270 & 8.85 & 0.653\\
    \midrule
    \multirow{3}{*}{$b{=}8$} & Nearest   & 0.311 & 8.14 & 0.670 \\
                             & Ours (no scene)    & \textbf{0.169} & \textbf{7.34} & \textbf{0.787} \\
                             & Ours (scene) & 0.174 & 7.53 & 0.779\\
    \midrule
    \multirow{3}{*}{$b{=}16$} & Nearest  & 0.271 & 7.66 & 0.768 \\
                             & Ours (no scene)    & \textbf{0.151} & \textbf{6.85} & \textbf{0.808} \\
                             & Ours (scene) & 0.155 & 7.00 & 0.801 \\
    \midrule
    $b{=}32$ (fit) & Ours (no scene) & 0.147 & 7.03 & 0.806 \\
                  & Ours (scene) & 0.151 & 7.16 & 0.799 \\
    \bottomrule
    \end{tabular}
\end{table}
\subsection{Class-Stratified Recovery}
\label{sec:class-recovery}
\begin{figure}
    \centering
    \includegraphics[width=0.49\linewidth]{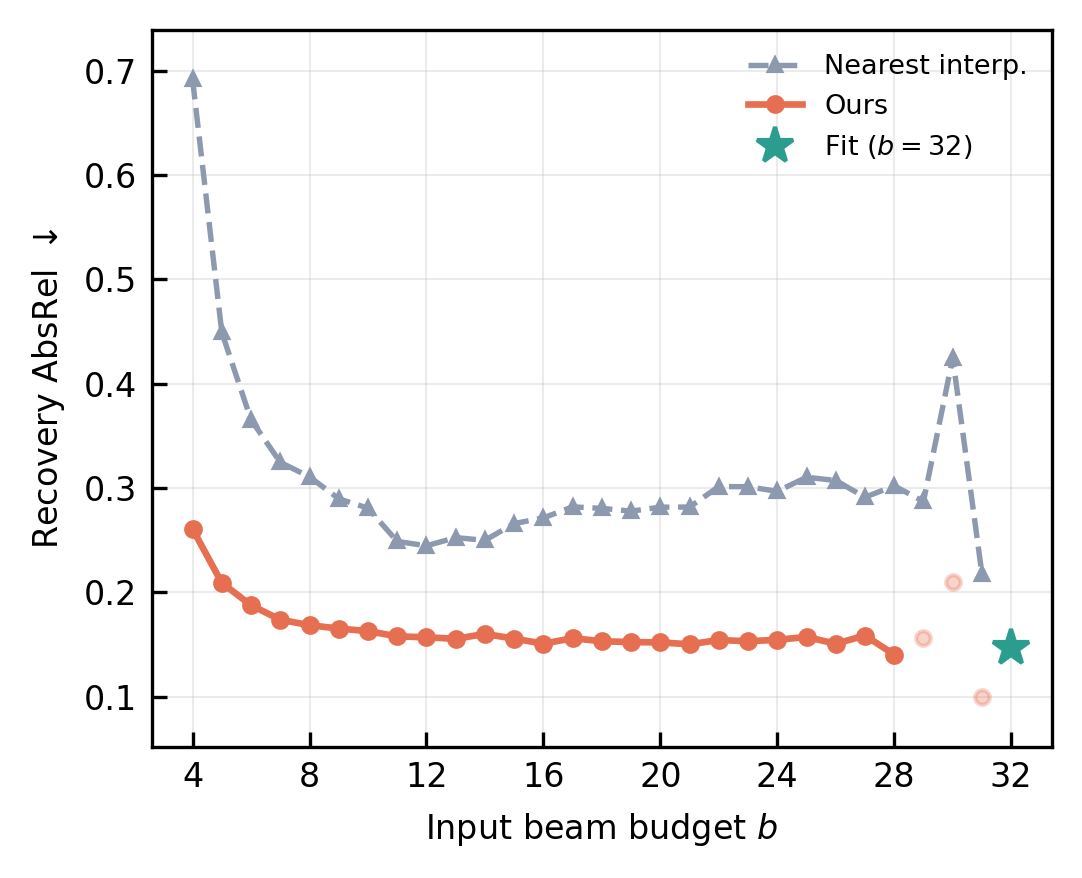}\hfill
    \includegraphics[width=0.49\linewidth]{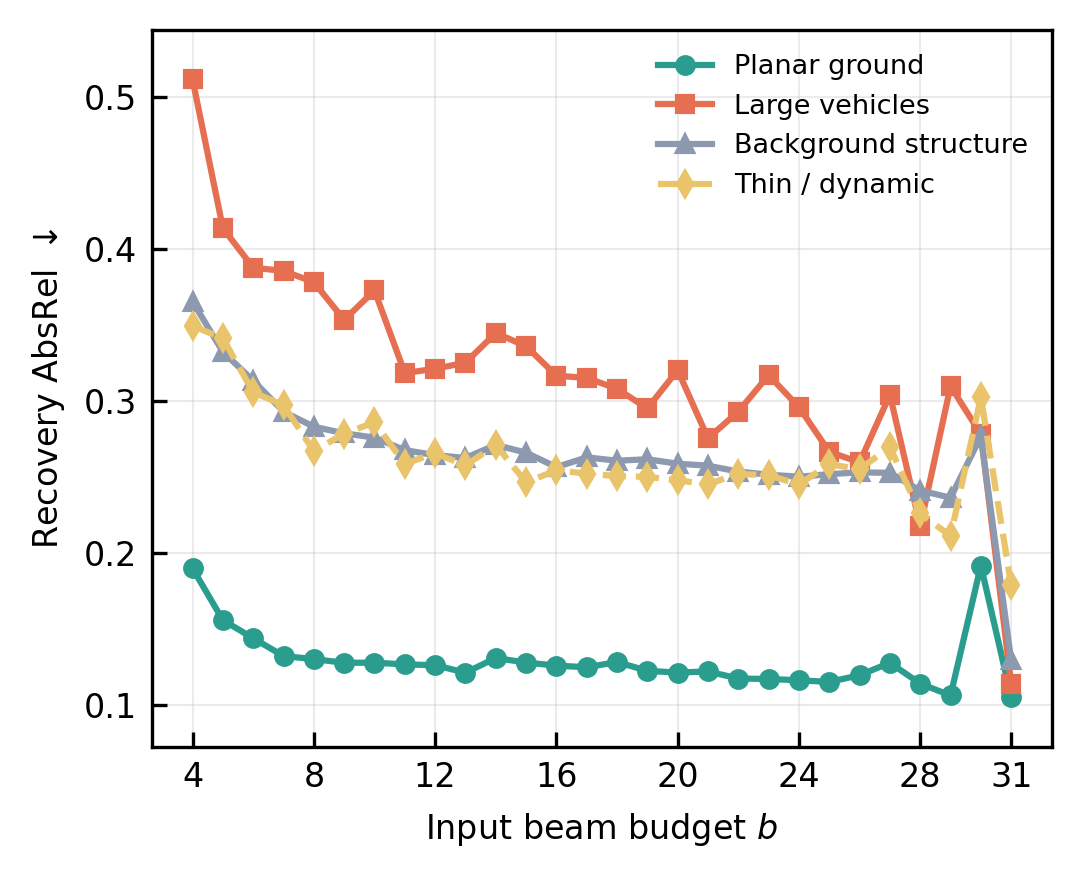}
    \caption{Recovery against input beam budget on nuScenes \cite{nuscenes} validation split. Left (overall recovery): the proposed model outperforms nearest-neighbor interpolation at every budget, with performance largely plateauing beyond $16$ beams. Right (recovery per semantic group): large-vehicle error (orange) falls steeply as additional beams constrain output geometry. Planar ground (green) shows marginal improvement.}
    \label{fig:class_curve}
\end{figure}

Recovery error is stratified by semantic class and tracked across beam budgets (\figurename~\ref{fig:class_curve} right). Recovery is strongly class-dependent and primarily driven by scene geometry rather than object category. Large planar surfaces (driveable surface, sidewalk, terrain) are reconstructed most accurately, consistent with the stronger geometric constraints provided by their locally smooth structure. Error increases for classes with dominant vertical extent and sharp depth boundaries (manmade, vegetation, barrier) and is highest for large vehicles (truck, trailer, bus, and construction vehicle). One possible explanation is that thin or irregular objects are more difficult to recover because sparse observations provide limited geometric constraints, leaving multiple depth configurations compatible with the observations. This interpretation is consistent with the observed budget dependence: large-vehicle error decreases rapidly as additional beams are retained, whereas planar surfaces show only marginal improvement. A partial vertical profile may be more compatible with multiple vehicle geometries, leaving unconstrained dimensions ambiguous until additional beams provide stronger geometric cues. Nevertheless, a residual error gap persists across all budgets, suggesting that depth discontinuities remain difficult to recover consistently across semantic classes. Thin and dynamic objects (pedestrian, bicycle, motorcycle) provide too few held-out beams per budget for reliable analysis. We therefore refrain from detailed interpretation but include them for completeness (dashed line). A more thorough analysis of output variance under severe decimation is left for future work.
\subsection{Effect of Scene Prompt}
\label{sec:prompt}
We compare recovery with and without scene-description conditioning to assess whether textual context aids reconstruction under severe beam sparsity. Scene conditioning provides no benefit for geometric recovery and marginally degrades it at every budget. This indicates that depth reconstruction depends primarily on the LiDAR conditioning rather than semantic scene context. Accordingly, the no-scene configuration is used as the primary setting for the reported results.
\section{Conclusion}
This work investigated camera-space LiDAR depth recovery under physically motivated beam decimation. We trained a single model across the full range of beam budgets and evaluated recovery exclusively on physically held-out LiDAR beams. The model outperforms nearest-neighbor interpolation at every budget, particularly in the sparsest regime ($\delta_{1.25}$ of $66.8$\% versus $45.1$\% at $b\!=\!4$), while recovery largely plateaus beyond $b\!=\!16$, approaching the performance obtained without decimation. Scene-description conditioning yielded no measurable benefit, suggesting that recovery is driven primarily by the LiDAR conditioning. The class-stratified analysis further shows that recovery quality depends strongly on the geometric structure. These results characterize the practical limits of camera-space depth recovery as LiDAR sampling density decreases and provide a basis for evaluating learned LiDAR enhancement under sensor sparsity.
\section*{Acknowledgments}
The research leading to these results is funded by the German Federal Ministry for Economic Affairs and Energy within the project “NXT GEN AI METHODS – Generative Methoden für Perzeption, Prädiktion und Planung" (grant no. 19A23014M).
\clearpage
\bibliographystyle{IEEEtran}
\bibliography{references}

\end{document}